# Anti-virus Autobots: Predicting More Infectious Virus Variants for Pandemic Prevention through Deep Learning

Glenda Tan Hui En[1*], Koay Tze Erhn[2*] and Shen Bingquan[3]

[1]Raffles Institution, Singapore
22yglen217g@student.ri.edu.sg
[2]Raffles Institution, Singapore
22ykoay421i@student.ri.edu.sg
[3]DSO National Laboratories, Singapore
SBingqua@dso.org.sg

## Abstract

More infectious virus variants can arise from rapid mutations in their proteins, creating new infection waves. These variants can evade one's immune system and infect vaccinated individuals, lowering vaccine efficacy. Hence, to improve vaccine design, this project proposes Optimus PPIme – a deep learning approach to predict future, more infectious variants from an existing virus (exemplified by SARS-CoV-2). The approach comprises an algorithm which acts as a "virus" attacking a host cell. To increase infectivity, the "virus" mutates to bind better to the host's receptor. 2 algorithms were attempted – greedy search and beam search. The strength of this variant-host binding was then assessed by a transformer network we developed, with a high accuracy of 90%. With both components, beam search eventually proposed more infectious variants. Therefore, this approach can potentially enable researchers to develop vaccines that provide protection against future infectious variants before they emerge, pre-empting outbreaks and saving lives.

## 1 Background and Purpose of Research Area

### 1.1 The Emergence of More Infectious Virus Variants

*Background and Motivation:* A small proportion of rapid mutations in viral genomes can significantly increase infectivity, leading to waves of new infections, which in turn create opportunities for more viral mutations to occur. This mechanism has prolonged the devastating pandemic caused by the novel severe acute respiratory syndrome coronavirus 2 (SARS-CoV-2). As variants of concern emerge, including Beta, Delta and Omicron [1], evidence points towards mutations in the SARS-CoV-2 spike glycoprotein that increase its binding affinity towards the human angiotensin-converting enzyme 2 (hACE2) receptor [2], thus raising transmissibility.

*Challenges and Objectives:* Currently, vaccines are the solution to reducing virus transmissibility by training one's immune system to develop a response against the virus. However, emergent virus variants can evade one's immune system and even infect vaccinated individuals, lowering vaccine efficacy. For instance, the Covid-19 Delta variant drastically lowered Pfizer BioNTech vaccine's efficacy from 93.7% to 39%, triggering global infection waves in 2021 [3, 4]. Unless vaccines are designed to combat both current and future more infectious variants, our pandemic battle will be a prolonged cat-and-mouse game where we struggle to develop booster shots to catch up with ever-mutating variants. Hence, we aim to use deep learning to predict future, more infectious virus variants. This can help researchers to prepare for vaccine production against these variants before they arise, pre-empting outbreaks and saving lives.

*Contributions:* In this paper, our contributions include:
   a. Developing a deep learning approach, Optimus PPIme, that generates mutations from an existing virus protein (exemplified by SARS-CoV-2) to predict future, more infectious variants.
   b. Developing a protein-protein interaction (PPI) transformer neural network, that can score the binding affinity between a virus protein and host receptor with a high test accuracy of 90%. Only protein primary sequences are needed as input.

In section 1, we introduce our Optimus PPIme approach and cite related work. Section 2 describes our research question and hypothesis while Section 3 documents the development of Optimus PPIme. Our results are shown in Section 4, while implications of our approach and future work are covered in Sections 5 and 6 respectively.

## 1.2  Our Deep Learning Approach – Optimus PPIme

Consider the following: A virus attacking a host cell aims to discover mutations that maximize its binding affinity to the host receptor, thereby increasing its infectivity. This is akin to a game character deciding on an optimal strategy to maximize its long-term reward –any action made at one time-step affects subsequent rewards. In the first context, our agent (the virus) can substitute any amino acid (AA) in its sequence of length L (L = 1273 for SARS-CoV-2) with 1 of the 20 AAs, giving rise to an action space of 20L. We exclude insertion and deletion mutations as these are less common in nature [5]. The environment (PPI transformer network) then outputs a PPI score for the proposed mutated protein (new state) and host receptor. The reward received is the change in PPI score (final – initial score of original virus protein $S_0$ and host receptor).

## 1.3  Related Work

PPIs are usually determined via tedious and costly high-throughput experimental methods, such as isothermal titration calorimetry and nuclear-magnetic resonance [6]. This necessitates the use of computational PPI models. However, many require 3D protein structures, which are harder to obtain than primary sequences. Even methods such as MuPIPR [7] –that only require primary sequences as inputs– fail to generalize to novel proteins. To address these, we propose a PPI transformer network that uses only primary sequences and generalizes to novel proteins.

Primary sequences can be represented as strings of letters denoting AAs. Protein motifs and domains (conserved functional patterns) [8] are also analogous to words and phrases. Furthermore, information is contained in primary sequences and natural sentences [9]. Such similarities make natural language processing (NLP) ideal for extracting protein features.

NLP tasks have seen state-of-the-art performance with the rise of transformers. These encoder-decoder networks adopt the self-attention mechanism, which relates different positions of a sequence to compute a rich representation of features [10]:

$$attention(Q,\ K,\ V) = softmax\left(\frac{QK^T}{\sqrt{d_K}}\right)V$$

Where Q, K, and V are the query, key and value matrices while $d_K$ is the dimension of K. Given that interactions between different AAs in a protein sequence give rise to the protein's structure and properties, transformer encoders are suitable protein feature extractor networks.

Lastly, we define a similarity measure between $S_0$ and a generated variant protein with sequence alignment scores derived from Block Substitution Matrix 62 (BLOSUM62) [11]. The BLOSUM distance between 2 sequences, $S = s_1...s_L$ and $S' = s'_1...s'_L$, is given by [12]:

$$D(S, S') = \sum_{i=1}^{L} (B_{s_i s_i} - B_{s_i s'_i})$$

## 2 Hypothesis of Research

Our research question is: Is a PPI predictor environment sufficient for virus agents to predict future, more infectious variants?

We hypothesize that given an environment that scores the binding affinity for a virus-host PPI with high accuracy, agents can predict future, more infectious variants.

## 3 Methodology

### 3.1 Dataset Collection

We train our PPI transformer on 31,869 experimentally-validated positive virus-human (V-H) interactions from BioGRID [13] and VirHostNet [14], excluding PPIs with SARS-CoV-2. We reference 34,836 negative V-H interactions from [15], giving rise to 66,705 training examples.

For the 50 novel virus protein test PPIs, we fix hACE2 as the second input protein. Positive PPIs include the original SARS-CoV-2 spike and its 23 variants listed in CoVariants [16], while 26 negative PPIs were sampled from unseen non-coronavirus viral proteins in [14]. All sequences were extracted from UniProt [17] and tokenized with Keras TextVectorization. We added start, end of sentence (SOS, EOS) and class (CLS) tokens, padding up to 1,300 tokens.

### 3.2 The Environment – PPI Transformer Network

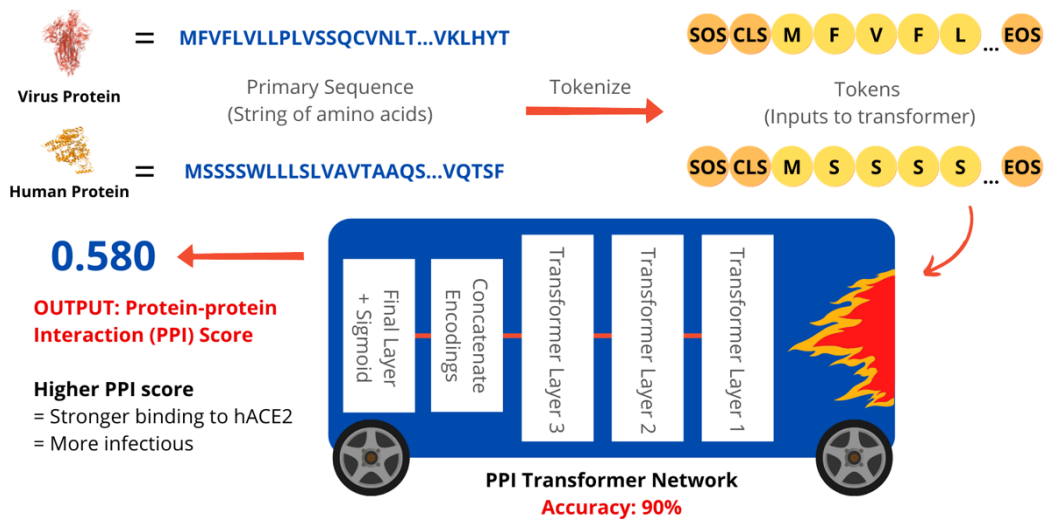

Figure 1. Overview of the PPI transformer network.

The inputs to our PPI transformer are the primary sequences of a virus and human receptor protein (represented by strings of amino acids). These strings of amino acids are then tokenized before they are fed to the transformer network. The transformer network then outputs a PPI score measuring how well the virus protein binds to the human receptor protein. The higher the PPI score, the stronger the virus-host binding and the more infectious the variant.

### 3.2.1 Experiment 1: Masked Language Modeling (MLM) Pre-training vs No MLM Pre-training

Initially, our transformer predicted the same value of 0.99 for all test data (see Appendix), indicating its inability to learn. This may be due to unbalanced dependencies and amplified output perturbations caused by layer normalization and aggravated by the Adam optimizer. Since learning semantics of input sequences improves downstream task performance (transfer learning) [18, 19], we pre-train our transformer on Masked Language Modeling (MLM) and fine-tune on PPI. In each sequence, we mask a random 15% of tokens with a mask token 90% of the time, or a random token otherwise. MLM attempts to reconstruct the original sequence.

### 3.2.2 Experiment 2: Sharpness-Aware Minimization (SAM) vs No SAM

Today's neural networks optimize on training loss, but this is inadequate in generalizing to unseen data (novel proteins for our PPI task) due to complex and non-convex training loss landscapes. SAM overcomes this by simultaneously minimizing loss value and loss sharpness. This algorithm enables a model to learn parameters whose entire neighbourhoods have uniformly low training loss values (low losses and curvature) rather than those with solely low loss values (low losses but complex curvature) [20, 21]. Hence, we determine the effects of SAM on generalizability. Using the model pre-trained with MLM in Experiment 1 as our baseline, 3 new models were trained with the addition of SAM on MLM pre-training only, PPI training only, and for both tasks (see Appendix for our implementation).

### 3.2.3 Experiment 3: Data Augmentation vs No Data Augmentation

Given that image augmentation improves image classifiers' robustness [22], we aim to determine if augmenting protein sequences could similarly boost our PPI test accuracies. The models were trained with 3 different augmentation techniques [23] during MLM pre-training: substituting a random AA with alanine –a generally unreactive AA (Alanine Sub)– or an AA of the highest BLOSUM62 similarity score –the most likely AA that can replace the original AA (Dict Sub), and reversing the whole protein sequence (Reverse). 25% of proteins were augmented and 20% of amino acid positions were replaced for substitution augmentations. Augmentation was not applied to PPI training as it distorts the proteins' structure and properties.

*Experimental Setup:* All PPI models adopted the same architecture (see Appendix) and were trained for the same number of epochs (50 for MLM pre-training, 15 for PPI training). They were evaluated on the same novel virus test set, with test accuracy and F1 scores as metrics.

## 3.3 The Agent – Proposing Future More Infectious Virus Variants

Initially, we attempted a Deep Q-Learning Network (DQN) agent (see Appendix). However, it required heavy computation and converged slowly due to our substantial search space of 20L (25,460 actions for SARS-CoV-2). Thus, we explore 2 more efficient algorithms, Greedy Search and Beam Search, to search for the variant with the highest PPI infectivity score. These algorithms rely on the greedy concept: making the optimal choice(s) at each time-step.

> **Greedy Search Algorithm**
>
> **Inputs:** current sequence S = original spike sequence $S_0$, actions A = 25460
> **while** BLOSUM distance <= 40 **do**:
>     Perform mutations on S within A to create a batch of variants with shape (25460, 1)
>     Tokenize, compute PPI scores and store in array P
>     Select the sequence with the highest PPI score, $S_{best}$ = argmax(P)
>     Update A = A – 20 actions for mutated position & S = $S_{best}$ if BLOSUM distance with $S_0 \leq 40$
> **return** S

Algorithm 1. Greedy Search algorithm.

> **Beam Search Algorithm**
>
> **Inputs:** S = {$S_0$}, A = 25460, beamwidth = 10, no. sequences with BLOSUM distance > 40, ($\eta$) = 0
> **while** $\eta$ < *beamwidth* **do**:
>     **for** *sequence s in S* **do**:
>         a.  Perform mutations on s to create a batch of variants with shape (25460, 1)
>         b.  Tokenize, compute PPI scores and store in array P
>     Select 10 best sequences with the highest PPI scores, $S_{best}$ = $argmax_{10}$(P)
>     Update S = {for s in $S_{best}$ if BLOSUM distance with $S_0 \leq 40$} and $\eta$ = 10 - length(S)
> **return** S

Algorithm 2. Beam Search algorithm.

We used Phyre2 [24] to predict the generated variant sequences' 3D structures. Then, possible binding modes of the variants and hACE2 were proposed by the HDOCK server, a hybrid of template-based modeling and free docking which achieved high performance in the Critical Assessment of Prediction of Interaction [25]. We used docking scores as a further metric to validate our proposed variants, where negative scores indicate spontaneous PPIs.

# 4 Results and Discussion

## 4.1 PPI Transformer Network Results

Table 1. Performance of the PPI models across all 3 transformer experiments.

| No. | MLM | SAM | Data Augmentation | Test Accuracy / % | Loss | F1 Score |
|---|---|---|---|---|---|---|
| 1 | ✓ | x | x | 44.0 | 1.080 | 0.417 |
| 2 | x | x | x | 50.0 | 4.288 | 0.658 |
| 3 | ✓ | MLM | x | 52.0 | 3.690 | 0.667 |
| 4 | ✓ | MLM + PPI | x | 72.0 | 0.707 | 0.774 |
| 5 | ✓ | PPI | x | 74.0 | 0.642 | 0.787 |
| 6 | ✓ | PPI | Reverse | 74.0 | 0.786 | 0.787 |
| 7 | ✓ | PPI | Dict Sub | 78.0 | 0.910 | 0.814 |
| **8** | **✓** | **PPI** | **Alanine Sub** | **90.0** | **0.438** | **0.906** |

*Experiment 1:* Although Model 2 (without MLM pre-training) achieved a higher test accuracy and F1 score than Model 1 (with MLM pre-training), it outputted the same PPI value for all test data (0.99), indicating its inability to learn. In contrast, MLM pre-training helped Model 1 to learn relevant protein features and it outputted different PPI scores for test data. Model 1 was thus used as the baseline for subsequent models to improve upon.

*Experiment 2:* MLM pre-training with SAM (Models 3 and 4) causes transformer layers that are nearer to the output to learn parameters which improve its predictions of the original AAs being masked (the MLM task). However, these MLM-specific parameters may not be best suited for our PPI task, which uses natural proteins without masking. Thus, SAM on PPI (Model 5) is essential for optimizing the parameters in our PPI task.

*Experiment 3:* Alanine Sub (Model 8) improved PPI test accuracy the most as it does not drastically alter the protein syntax as compared to Reverse and Dict Sub. This is likely due to alanine's nonpolar methyl side-chain (see Appendix), giving rise to alanine's unreactivity [26]. Model 8 was therefore chosen as the optimal PPI transformer network environment.

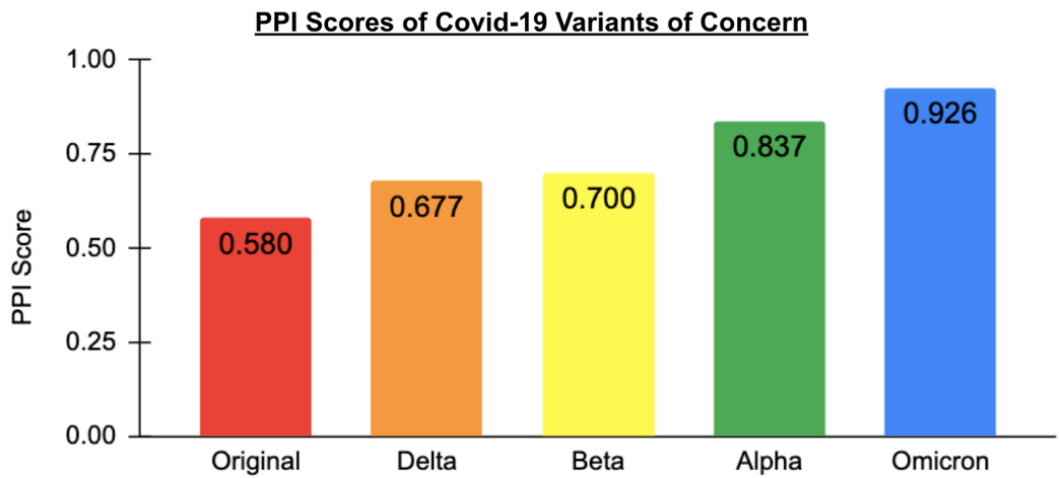

Figure 2. Graph showing PPI scores for different Covid-19 variants of concern.

All variants of concern achieved higher PPI scores than the original spike protein (0.580). The Delta variant (0.677) achieved a higher PPI score than the original, although lower than Alpha (0.837) and Omicron (0.926). These results reflect that our PPI transformer network can make real-world predictions which corroborate well with current Covid-19 research data [27, 28].

## 4.2  Virus Agent Algorithm Results

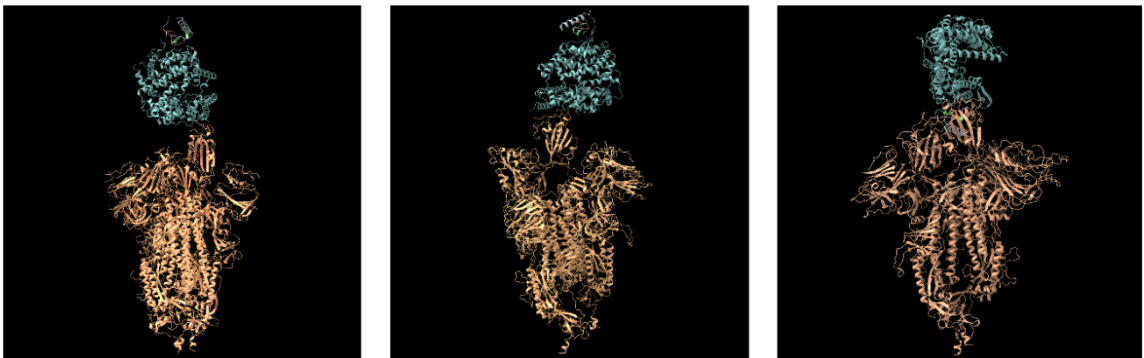

Figure 3. 3D structures of original, greedy and beam search spike proteins bound to hACE2.
Legend: blue: hACE2 receptor, orange: spike glycoprotein.

Table 2. hACE2 binding metrics for spike protein variants generated by greedy & beam search.

| Algorithm | PPI Score | Docking Score | RMSD with $S_0$ / Å |
|---|---|---|---|
| Greedy Search | 0.99969 | –214.76 | 0.870 |
| Beam Search | 0.99973 | –218.90 | 1.095 |

Based on Table 2, the spike variant proposed by Beam Search attained higher PPI and more negative docking scores than that for Greedy Search, reflecting its greater hACE2 binding affinity. Unlike greedy search which always exploits the best immediate action, beam search strikes a balance between exploitation and exploration (trying a new action). Since AAs in a sequence interact with one another, by considering 10 sequences at each time-step, beam search is likelier to find mutations that may not maximize short-term rewards but will optimize long-term rewards due to future AA interactions. From Figure 3, the proposed variants' structures also have little deviation from the original protein, with RMSDs close to those of current variants (see Appendix). Therefore, an agent armed with a PPI transformer network can propose future more infectious variants, proving our hypothesis. The variants can then be validated experimentally.

## 5   Implications and Conclusion

We discovered that given an accurate PPI transformer network that measures the infectivity of a proposed variant, our Optimus PPIme approach can effectively predict possible more infectious variants. This narrows the scope of mutated virus proteins for docking and wet-lab testing to validate the variants' infectivity and feasibility.

With only knowledge of the virus and receptor protein sequences, our Optimus PPIme approach can be applied to other dangerous viruses to expedite vaccine development before infectious variants become widespread.

## 6   Limitations and Future Work

Currently, our Optimus PPIme approach does not consider insertion or deletion mutations in the virus protein, which are also likely to occur in nature. Besides that, the ability of a virus to evade vaccine antibodies is another metric for infectivity, which we did not consider in our Optimus PPIme approach.

Hence, future work can be done to generate insertion or deletion mutations in the virus variant, and to use the evasion of antibodies as a further metric for infectivity.

## 7   Acknowledgements

We would like to thank our mentor, Dr Shen, for his invaluable guidance and advice throughout this research project!

# 9 Appendix

## 9.1 PPI Dataset Breakdown

Table 3. Breakdown of train and test datasets for the PPI transformer network.

| Dataset | Train | Test |
|---|---|---|
| BioGRID | 11,612 (+) | N/A |
| VirHostNet | 20,257 (+) | 26 (–) |
| Kshirsagar et al. | 34,836 (–) | N/A |
| CoVariants | N/A | 24 (+) |
| Total | 66,705 | 50 |

In order for our PPI transformer to generalize to unseen virus proteins and make unbiased predictions of future virus variants' infectivity, the training set does not contain V-H interactions involving SARS-CoV-2. Instead, only the PPI transformer's test set contains V-H interactions involving Covid-19 variants and hACE2.

## 9.2 Our SAM Implementation

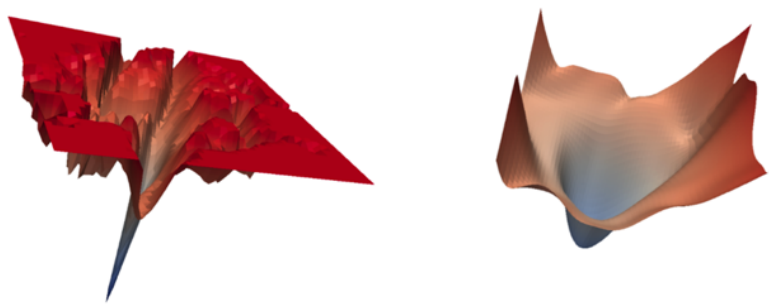

Figure 4. Loss landscapes without SAM training (left) and with SAM training (right) [20].

**PPI SAM Algorithm**

**Inputs:** neighbourhood ($\rho$) = 0.05, PPI model's weights $w_0$, timestep t = 0
**while** *not converged* **do**
1. Sample batch B = $\{(x_1, y_1), \ldots(x_8, y_8)\}$, where x = [$V_{token}$, $H_{token}$]
2. Backpropagation 1: Compute training loss and gradient g
3. Scale gradient by factor $\rho$ / (|| g || + 1e-12) and update weights
4. Backpropagation 2: Compute training loss and final gradient G
5. Update weights: $w_{t+1} = w_t - \alpha * G$
   t = t + 1
**return** $w_t$

Algorithm 3. PPI SAM algorithm.

From Figure 4, a loss landscape without SAM training has a sharp global minimum (left) and is difficult to converge, whereas SAM training results in a loss landscape with a wide minimum (right) that is easier to converge [20]. From Algorithm 3, our SAM implementation involves 2 backpropagation steps with a scaling step between them.

### 9.3 Visualization of Protein Augmentation Techniques

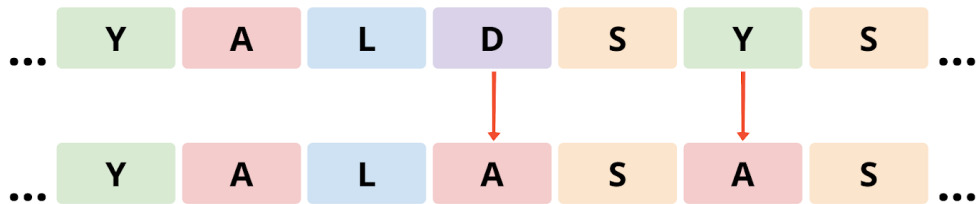

Figure 5a. Visualization of Alanine Sub augmentation.

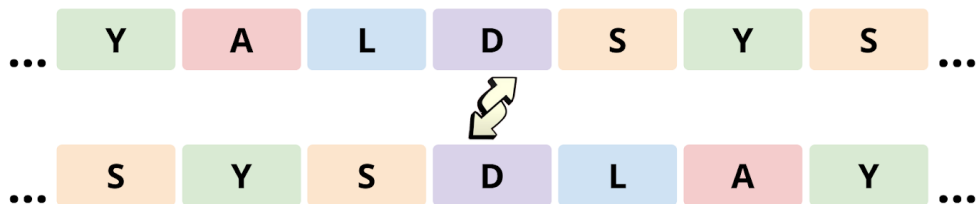

Figure 5b. Visualization of Reverse augmentation.

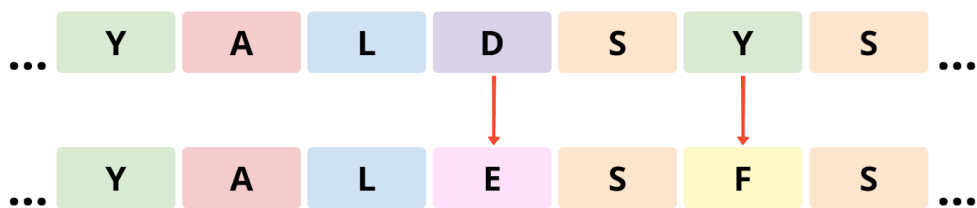

Figure 5c. Visualization of Dict Sub augmentation.

The augmentation techniques were implemented by performing string manipulations on the protein sequence and subsequently feeding the augmented proteins to the transformer network during training. Combinations of augmentations (e.g. all 3, Alanine Sub & Reverse, Alanine Sub & Dictionary Sub, Reverse & Dictionary Sub) were also attempted but they produced models with lower accuracies.

### 9.4 Skeletal Structure of Amino Acid Alanine

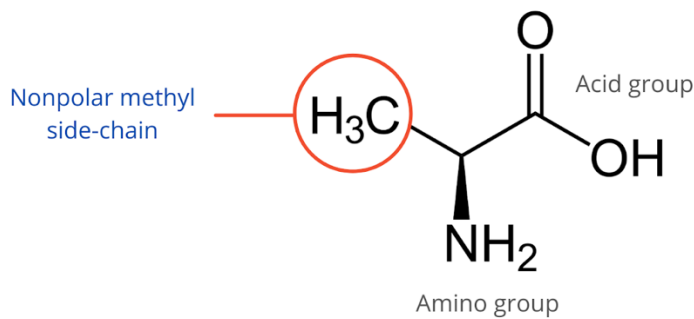

Figure 6. Skeletal structure of alanine with its nonpolar methyl side-chain.

Alanine is a generally unreactive amino acid, attributed to its nonpolar methyl side-chain. This makes Alanine Substitution a suitable candidate for data augmentation.

### 9.5 Model Architectures

#### 9.5.1 Architecture of the PPI Transformer Network

The PPI transformer network was implemented using the tf.keras framework. The network consists of an encoder to extract features of the virus and receptor proteins, a Concatenate layer to concatenate the encodings of the 2 proteins and finally a 1-unit Dense layer with sigmoid activation which outputs the PPI score. No decoder is used as only an encoder is required for extracting the protein features.

**Encoder:** The encoder is adapted from the Bidirectional Encoder Representations from Transformers (BERT) encoder and it comprises 3 identical layers. Each layer is made up of a MultiHeadAttention component and a fully-connected feed-forward network. A residual connection is applied around the 2 components, followed by LayerNormalization.

**Hyperparameters:** After experimenting with different values for the hyperparameters, we arrive at this combination of hyperparameters which produced the best-accuracy model of 90%.

Table 4. Hyperparameters for the PPI transformer encoder.

| Hyperparameter | Value |
| --- | --- |
| Number of transformer layers in encoder | 3 |
| Number of heads for MultiHeadAttention | 8 |
| Number of embedding dimensions | 128 |
| Dropout rate | 0.1 |
| Adam Optimizer learning rate | 0.001 |

| Batch size | 8 |
|---|---|
| Maximum length of proteins | 1,300 |
| MLM pre-training epochs | 50 |
| PPI training epochs | 15 |

### 9.5.2 Architecture of the DQN Agent

The inefficient DQN agent adopts the same encoder architecture as the PPI transformer network, and the encoder is connected to a 25,460-unit Dense layer with softmax activation. 25,460 represents the number of possible mutations the SARS-CoV-2 spike protein can perform at each time-step (1,273 AA positions x 20 amino acids). The DQN agent was trained for 1000 episodes with a learning rate of 0.7, a batch size of 128 and a minimum replay size of 500.

## 9.6 Failed Results

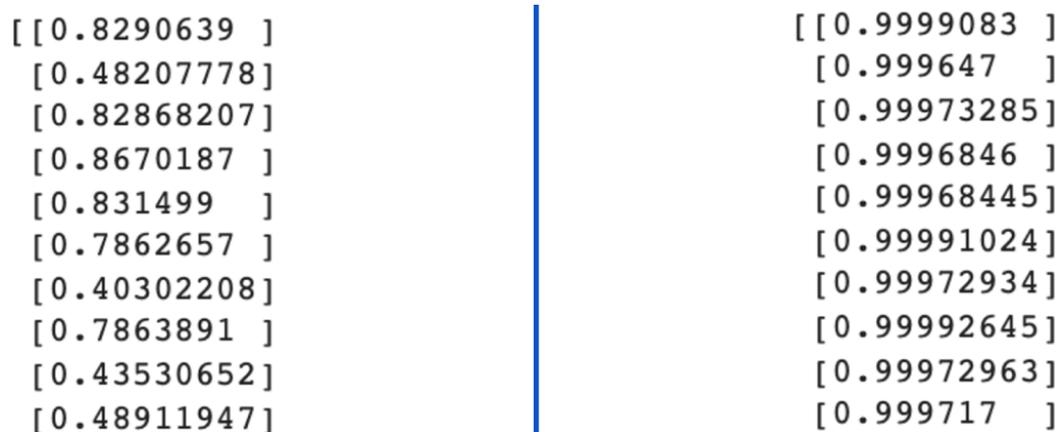

Figure 7. (from left to right) PPI scores outputted by Model 1 (with MLM pre-training) and Model 2 (without MLM pre-training).

```
0 , FR: -0.0024803877 , FP: 0.5405822        Updating target network weights
Updating target network weights              16/16 [==============================]
1 , FR: -0.0040413737 , FP: 0.5390212        16/16 [==============================]
Updating target network weights              16/16 [==============================]
2 , FR: 0.0059351325 , FP: 0.5489977         819 , FR: 0.0035093427 , FP: 0.5465719
Updating target network weights              Updating target network weights
3 , FR: 0.00032663345 , FP: 0.5433892        16/16 [==============================]
Updating target network weights              16/16 [==============================]
4 , FR: 0.0019819736 , FP: 0.54504454        16/16 [==============================]
Updating target network weights              820 , FR: 0.0015875101 , FP: 0.5446501
                                             Updating target network weights
                                             16/16 [==============================]
                                             16/16 [==============================]
                                             16/16 [==============================]
                                             821 , FR: -0.0005326867 , FP: 0.5425299
                                             Updating target network weights
                                             16/16 [==============================]
                                             16/16 [==============================]
                                             16/16 [==============================]
                                             822 , FR: -0.0005326867 , FP: 0.5425299
                                             Updating target network weights
                                             16/16 [==============================]
                                             16/16 [==============================]
                                             16/16 [==============================]
                                             823 , FR: -0.0011529922 , FP: 0.5419096
                                             Updating target network weights
```

Figure 8. (from left to right) First and last 5 episodes of DQN training.
Legend: FR: final reward (final - initial PPI score of $S_0$ and hACE2) at the end of each episode,
FP: final PPI score at the end of each episode.

From Figure 7, Model 2 outputs the same value of 0.99, indicating its inability to learn. In contrast, MLM pre-training helps Model 1 to output different PPI values, showing that the model has successfully learned the features of proteins. From Figure 8, DQN's final reward does not increase and oscillates between being positive and negative, reflecting the agent's inability to converge.

### 9.7 Confusion Matrices of the PPI Transformer Networks

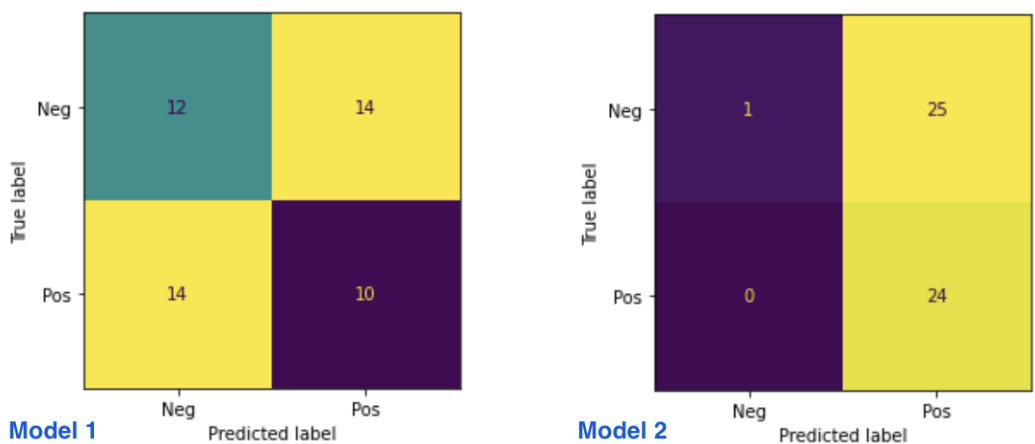

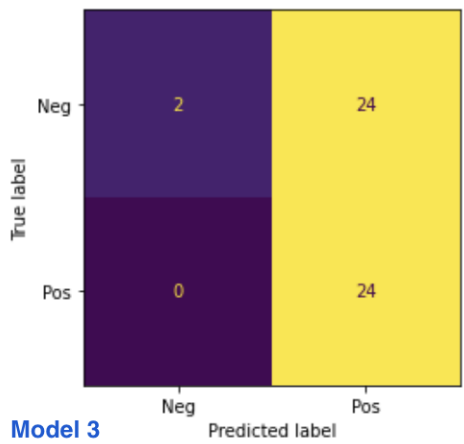
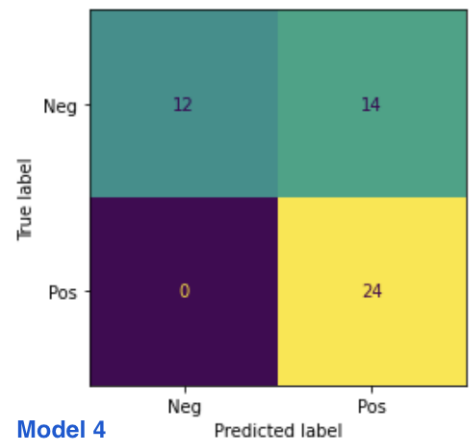
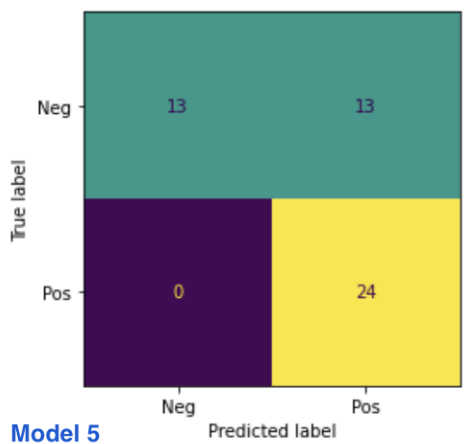
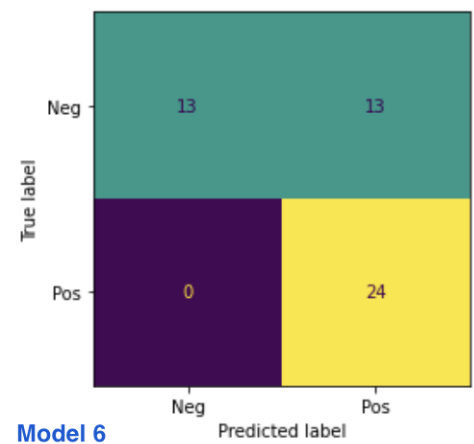
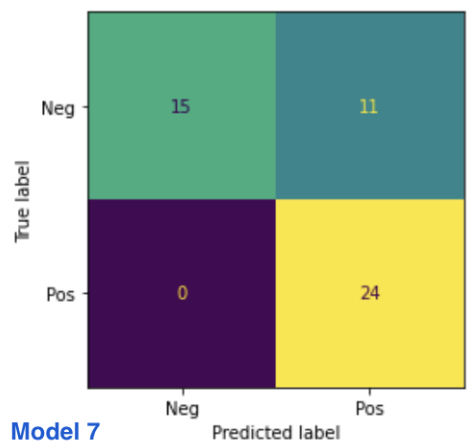
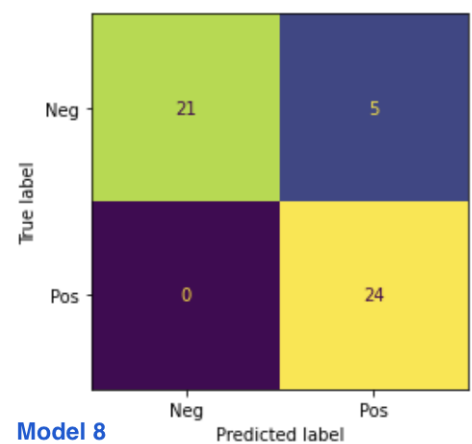

Figure 9. Confusion matrices of PPI transformer networks 1 to 8.

Based on the confusion matrices, Model 8 achieved the highest test accuracy. It correctly predicted all 24 positive interactions and 21 out of 26 negative interactions in our test set.

## 9.8 Comparison of Our Proposed Variants with Covid-19 Variants of Concern

Table 5: Number of spike mutations and RMSDs of the different variants.

| Variant | Number of Spike Mutations | RMSD with $S_0$ / Å |
|---|---|---|
| Alpha | 4 | 0.897 |
| Beta | 8 | 0.901 |
| Delta | 8 | 1.091 |
| Omicron | 17 | NA |
| Greedy (Proposed) | 6 | 0.870 |
| Beam (Proposed) | 6 | 1.095 |

The RMSD values were derived by matchmaking the variants' Protein Data Bank (PDB) files with the original spike protein in ChimeraX. At the time of submission, Omicron's PDB file and hence its RMSD were not yet available due to the variant's recency.